\documentclass{article}
\usepackage{acra}
\usepackage{soul}

\usepackage{graphicx}
\usepackage{subfig}
\usepackage{placeins}
\usepackage[inkscapearea=page]{svg}

\usepackage{enumitem}
\usepackage{url}
\usepackage{multirow}

\usepackage{flushend}
\usepackage{balance}
\usepackage{microtype}
\usepackage[UKenglish]{babel}
\usepackage{pdfcomment} 
\newsavebox\mybox

\usepackage{booktabs}

\title{Collaborative Object Handover in a Robot Crafting Assistant}
\author{
Leimin Tian \\ CSIRO Robotics, Australia \\ 
leimin.tian@csiro.au
\And
Shiyu Xu \and Kerry He \and Rachel Love \\ Monash University, Australia \\ 
{shiyu.xu, kerry.he, rachel.love}@monash.edu
\AND
Akansel Cosgun \\ Deakin University, Australia \\ 
akan.cosgun@deakin.edu.au
\And
Dana Kuli\'{c} \\ Monash University \& CSIRO Robotics, Australia \\ 
dana.kulic@monash.edu
}

\begin{document}

\maketitle

\begin{abstract}
Robots are increasingly working alongside people, delivering food to patrons in restaurants or helping workers on assembly lines. These scenarios often involve object handovers between the person and the robot. To achieve safe and efficient human-robot collaboration (HRC), it is important to incorporate human context in a robot's handover strategies. We develop a collaborative handover model trained on human teleoperation data collected in a naturalistic crafting task. To evaluate its performance, we conduct cross-validation experiments on the training dataset as well as a user study in the same HRC crafting task. The handover episodes and user perceptions of the autonomous handover policy were compared with those of the human teleoperated handovers. While the cross-validation experiment and user study indicate that the autonomous policy successfully achieved collaborative handovers, the comparison with human teleoperation revealed avenues for further improvements.
\end{abstract}

\section{Introduction}\label{sec:intro}
Object handover is a common task in human-robot collaboration (HRC), in which an item is transferred between a human and a robot~\cite{ortenzi2021object}. To develop an autonomous human-robot handover policy, existing approaches either react to pre-defined triggers (e.g., specific hand gestures~\cite{kwan2020gesture}), or train machine learning models using data collected from isolated handover episodes between two people or in human-robot pairs (e.g.,~\cite{chan2021experimental,wiederhold2024hoh}). However, using pre-defined triggers imposes learning costs for the person and interrupts the flow of activities, requiring additional physical and mental effort for the person to explicitly command the robot~\cite{carissoli2023mental}. Further, as handovers are context-dependent~\cite{ortenzi2021object}, training from isolated episodes limits the resulting model's applicability in handovers that occur in naturalistic contexts with pre- and post-handover tasks. 

There has been growing interest in the social-interactive aspects of handovers in addition to the control and trajectory planning aspects, focusing on how a person and a robot may communicate and coordinate during handovers~\cite{moon2021design,iori2023dmp}. While these studies investigated communicating information such as intention or hesitation during individual episodes of object transfer, how people and robots collaborate in a complex task involving multiple handovers remains largely under explored. Handovers in a broader task context require providing objects at the right place and the right time, when the person may be engaged in the task and not giving explicit cues to the robot. Further, research on human behaviours in handovers largely focuses on hand locations, while additional informative behaviours such as gaze or whole body movements are under explored~\cite{hart2014gesture}. Evaluation of human-robot handover policies is often conducted in a short-term, handover-centred design without incorporating human and task contexts, e.g., focusing on the smoothness of the robot's motions and trajectories~\cite{duan2024human}. This results in a gap between a handover policy's outcomes in isolated episodes compared to when multiple handovers are serving an overall task goal. For example, a policy that emphasises optimising the instantaneous ergonomics risk factors in individual handovers may become over-assistive and reduce worker well-being in repeated interactions~\cite{zolotas2024imposing}.

\begin{figure*}[tb]
  \centering
    \pdftooltip{\includegraphics[width=\linewidth]{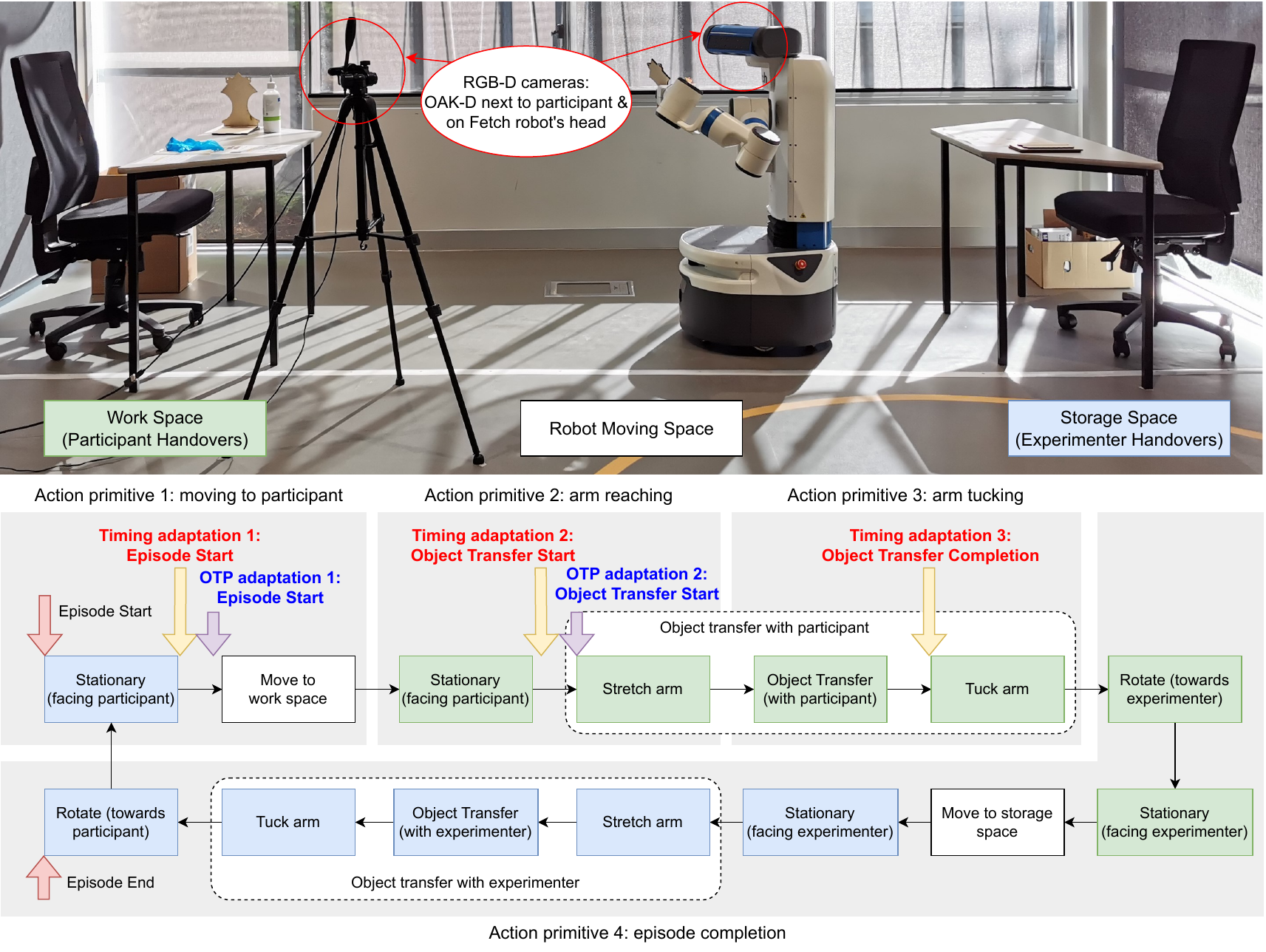}}{Top half of the image shows a photo of the experimental layout, with the work space where participants sit on the left and storage space where the experimenter sits on the right. The Fetch robot is moving between these two desks. A partially built birdhouse can be seen on top of the work desk. A tripod with a camera on top stands next to the work desk. A storage box can be seen next to the storage space. The robot also has a camera on its head. Below the photo, the state transitions during a handover episode are shown, with three timing adaptations and two handover location adaptations highlighted at key state transitions.} 
  \caption{Our autonomous handover policy performs temporal adaptations at episode start, object transfer start, and object transfer completion. Spatial adaptation of object transfer position (left-hand side of the participant, middle of the work space, right-hand side of the participant) is performed at episode and object transfer starts.}
  \label{episode_overview}
\end{figure*}

Therefore, we are motivated to understand human-robot handovers in a naturalistic task context where they serve a collaborative goal. In particular, we develop an autonomous handover policy, as shown in Figure~\ref{episode_overview}, trained with the Functional And Creative Tasks Human-Robot Collaboration dataset (the FACT HRC dataset)~\cite{tian2023crafting}. In the FACT HRC dataset, a mobile manipulator robot was teleoperated by a hidden human operator to serve as a crafting assistant to 20 users, each in an hour-long interaction session that involved multiple handovers of diverse objects used for the crafting task. We then replicate the crafting assistant task to evaluate the autonomous handover model with 20 new participants. Further, by comparing the autonomous handover policy and the human operator's handover strategies, we investigate design implications to improve collaborative handovers. Our main contributions are:
\begin{enumerate}
    \item We developed a collaborative handover policy for a mobile manipulator robot, which adapts the handover timing and object transfer location based on a user's upper body movements;
    \item We conducted a user study and evaluated the developed autonomous policy during multiple purposeful handovers with mixed robot roles in a collaborative task context;
    \item We compared the objective performance and subjective user perception of autonomous and teleoperated handovers to inform HRC research;
    \item The code and data are open-sourced\footnote{\url{https://doi.org/10.26180/25449640}}.
\end{enumerate}


\section{Related Work}\label{sec:bg}
As reviewed by Duan~et~al.~\cite{duan2024human}, research on complex long-sequence tasks with handover policies capable of spatial-temporal collaboration is extremely limited. Recent handover research investigates how different methods for trajectory generation, motion planning, or timing control influence the objective handover performance (e.g., duration or success rate) and subjective human perceptions (e.g., trust or fluency). Existing studies focus on human-to-robot (H2R) or robot-to-human (R2H) handovers, with limited research on bidirectional handovers, i.e., exchanging objects. In H2R handovers, previous work focused on predicting human behaviours or intentions under uncertainty with machine learning models (e.g.,~\cite{mavsar2022rovernet,yang2022model}). In R2H handovers, previous work focused on enhancing human comfort and increasing the movement efficiency of both parties~\cite{qin2022task,lagomarsino2023maximising}. 

Existing handover policies may adopt learning-based, control-based, or analysis-based approaches. For example, K{\"a}ppler~et~al.~\cite{kappler2023optimizing} developed an adaptive method for a table-mounted robot arm, which updated the object transfer location based on the location of a user's hand. This adaptive R2H handover model was compared with a non-adaptive model using a pre-defined object transfer location in four repeated handovers. They found that participants exhibited motor learning and adaptation to the robot's handover, which necessitated evaluating handover models in repeated episodes. 
Kedia~et~al.~\cite{kedia2024interact} developed a transformer-based human intent prediction model pre-trained on human-human collaborative manipulation data and fine-tuned on human-robot data with a teleoperated robot arm. They incorporated task context, such as a person and the fixed robot arm taking turns to each pick up one of two objects from a cart to place onto a shared table. However, this work focused on short interaction (3-15s) without direct object transfers between the human and the robot.
Zhuang~et~el.~\cite{zhuang2022goferbot} developed an R2H handover model in which a table-mounted robot arm delivered four legs of a table to a human to assist in assembly. They evaluated handover time, success rate, action recognition performance, and participants' perceptions. This work incorporated collaborative task context in repeated handovers. However, their approach focused on predicting the handover initiation timing, leaving out other timing and location coordination in the whole handover process.

As identified by Ortenzi~et~al.~\cite{ortenzi2021object}, there is limited work on adaptive handovers incorporating human social and communicative cues beyond hand locations. Moreover, pre- and post-handover tasks and HRC context were rarely considered in current handover research. Our previous work~\cite{tian2023crafting} investigated spatial-temporal adaptation in a mix of H2R, R2H, and bidirectional handovers contextualised in an hour-long HRC task of assembling and painting a wooden birdhouse. While the mobile manipulator robot was teleoperated by a hidden human operator (i.e., Wizard-of-Oz), the FACT HRC dataset collected in this work facilitates the development of a collaborative handover policy by learning from the human operator's strategies.

\section{Methodology}\label{sec:method}
To develop a collaborative handover policy capable of spatial and temporal adaptation based on human and task context in purposeful handovers, we used data from a human teleoperator's adaptive handovers during a crafting task, where multiple purposeful handovers of mixed H2R, R2H, and bidirectional handovers are required~\cite{tian2023crafting}, as further described in Section~\ref{subsec:method-ml}. We then evaluate the autonomous handover policy in the same HRC task to understand the task outcomes and users' subjective experience, as described in Section~\ref{subsec:method-s2}. We used the Fetch mobile manipulator robot~\cite{wise2016fetch} with a 7-DOF arm and a differential drive base, with the robot holding a basket for object delivery and return during handovers. Implementation of the handover model and data collected from our user study (ROS bags, processed CSV files, anonymous questionnaire responses) are open-sourced for research purposes. The study protocols were reviewed and approved by Human Research Ethics Committee at Monash University (Project ID 31927).

\subsection{Collaborative handover model}\label{subsec:method-ml}
We trained the autonomous handover policy using the FACT HRC dataset~\cite{tian2023crafting}, which contains 20 HRC sessions (10 women, 10 men, age 27.0$\pm$5.2, 19 right-handed, 1 left-handed) with 565 handover episodes (57\% R2H, 34\% H2R, 9\% bidirectional) in total. Each HRC session is approximately an hour long with multimodal data collected at time steps of 0.1s. We used the FACT-processed segment of the dataset, which includes non-identifiable CSV data of the robot's status (velocity and coordinates of the base and arm joints, coordinates of the end effector goal for object transfers), the operator's controls, the facial and upper body keypoints (normalised $(x,y,z)$ coordinates of 25 skeletal keypoints with prediction confidences~\cite{bazarevsky2020blazepose}) of participants estimated from an RGB-D camera (OAK-D) positioned next to the participants, and emotion estimations (categorical emotions with intensity, arousal and valence values, inferred from participant's facial expressions~\cite{toisoul2021estimation}) from this camera and the robot's onboard camera. Figure~\ref{episode_overview} provides an overview of the states in each handover episode in the FACT HRC dataset. Within an episode, the human operator was observed to focus their temporal and spatial adaptation, i.e., when and where to perform certain handover actions, at key time points. Thus, instead of continuously predicting the robot's base and arm actions at every time step, which incurs high computational cost and latency, we segmented the handover episode into four action primitives (APs). In Section~\ref{subsec:ml-primitive}, we discuss our validation of this discretisation of the handover adaptation strategies.

We trained supervised learning policies with the operator's actions as the ground-truth labels. As illustrated in Figure~\ref{episode_overview}, to achieve temporal adaptation, we trained binary classifiers to predict episode starts (i.e., when the robot starts moving its base towards the participant during AP1), object transfer starts (i.e., when the robot starts stretching out its arm to initiate object transfers during AP2), and handover completions (i.e., when the robot starts tucking its arm and concluding the current handover episode with the participant during AP3). During training, we excluded object transfers with the experimenter (AP4) and focused on learning the operator's strategies when collaborating with na\"ive participants. To achieve spatial adaptation, we trained a 3-way classifier to predict the object transfer positions (OTPs) as chosen from the default left (from the robot's perspective, i.e., to the right-hand side of the participant), middle (centre of the work space), or right (left-hand side of the participant) handover goal locations implemented in~\cite{tian2023crafting}. The operator chose one of these OTPs to ensure an object can be passed at a location convenient to the participant and suitable for the collaborative task context. Note that the OTP adaptation was performed at both episode starts during AP1 (initial estimation) and object transfer starts during AP (updated estimation) as informed by the operator's strategies observed in~\cite{tian2023crafting}. In Section~\ref{sec:ml}, we further discuss our machine learning experiments testing different features and model structures.
We implemented the machine learning models using the Python TensorFlow library~\cite{tensorflow2015-whitepaper}. The trained classifiers were incorporated with the ROS-based teleoperation framework developed in~\cite{tian2023crafting}. Specifically, the classifier outputs determined when and which pre-recorded APs the robot would execute, replacing the operator's controls.

\subsection{User study evaluation}\label{subsec:method-s2}
We replicated the HRC task in~\cite{tian2023crafting}, where the Fetch mobile manipulator robot served as a crafting assistant by transporting a list of task-relevant objects between a participant who sat at a work desk and an experimenter who sat at a storage desk (Figure~\ref{episode_overview}). The hour-long session was divided into three stages, namely Stage 1 (Preparation, where the participant receives protective equipment, such as a box of tissues and gloves), Stage 2 (Assembly, where the participant receives wooden pieces and glues them together into a birdhouse), and Stage 3 (Painting, where the participant receives brushes and paints to colour the birdhouse to their liking). The list of objects and crafting instructions were displayed on the work and storage desks. The participants were instructed to return objects they no longer need to necessitate a mix of R2H, H2R, and bidirectional handovers. 

An observer annotated the handover episodes' quality (Good, Bad, Neutral) and type (H2R, R2H, bidirectional), took observation notes, and monitored for program execution and technical issues. ROS bags and processed CSV data in the same format as the FACT HRC dataset were collected.
The participant filled in a questionnaire, which collected their demographic information and ratings on the robot self-efficacy scale~\cite{robinson2020robot} in the pre-study section. After each of the three stages, the participants rated their subjective perception of the robot~\cite{bartneck2009measurement} and the HRC~\cite{hoffman2019evaluating}. The robot was paused when the participants filled in the questionnaire. After the whole HRC session, the participants were asked to guess whether the robot was autonomous or teleoperated. They were also interviewed to gather further understanding of their behaviours and experiences.

We conducted the experiments at the Monash Robotics lab. We recruited 20 participants (9 women, 11 men, age 26.3$\pm$4.6, 19 right-handed and 1 left-handed) from the university student and staff population via printed and digital advertisement, direct contact and snowballing. Participants who took part in our previous experiments~\cite{tian2023crafting} were excluded from this study. In addition to analysing objective and subjective performance of the handover policy in Section~\ref{sec:s2} as a within-subject study, we compared the autonomous handovers with the teleoperated handovers in Section~\ref{sec:S1vsS2} as a between-subject study.
We hypothesise that the autonomous handover policy will achieve objective and subjective performance similar to human teleoperation.

\section{Machine Learning Experiments}\label{sec:ml}
We compared different features, model structures, and training approaches in cross-validation (CV) experiments to identify a suitable implementation of the autonomous handover policy. Here we report performance of the classifiers chosen for the user study.

\subsection{FACT HRC dataset evaluation}\label{subsec:ml_picked}
To predict base and arm actions during handovers, we tested features based on the emotion predictions from both cameras and features based on the upper body poses inferred from the OAK-D camera. While combining emotions and poses yielded better performance than poses alone, this improvement was limited due to the low accuracy of the emotion inference. Further, the human operator reported using emotions mainly for evaluating handover quality rather than determining handover strategies~\cite{tian2023crafting}. Considering the increased latency of extracting both emotion and pose features from two camera feeds, we chose the pose features, i.e., ($x$, $y$, $z$, confidence) of 25 upper body keypoints, as inputs of the autonomous handover model.

As shown in Figure~\ref{episode_overview}, handover adaptation happened mainly during a subset of state transitions in the whole episode. Thus, when predicting arm and base actions in terms of timing adaptation (episode start, OTP start, OTP completion), we first extracted all the key time steps where the arm and base status changed based on the robot status data in the FACT HRC dataset. We then labelled data within a 5s window (50 steps, duration of the shortest state, i.e., stretch/tuck arm) before each episode start, object transfer start, and object transfer completion as \textit{True} while all other time steps were labelled as \textit{False}. This transformed continuous arm and base action prediction into three binary timing classifications. To balance \textit{True} and \textit{False} instances in each binary classification, we down-sampled the time steps labelled as \textit{False} randomly.

Similarly, for predicting spatial adaptation of OTP, we used a subset of the data within 5s of episode starts and object transfer starts. As the OTP classes were unbalanced in the FACT HRC dataset (33\% left, 65\% middle, 2\% right), we over-sampled the training data by adding synthesised training data of left and right OTP. We first synthesised left OTP data by generating random values on a Gaussian constructed with the mean and standard deviation of original left OTP feature values, doubling the amount of left OTP data. We then synthesised right OTP data by mirroring the feature values of symmetrical skeletal keypoints of the left OTP (original and synthesised), e.g., interchanging the ($x$,$y$,$z$,confidence) of the left and right shoulder keypoints.

In terms of model structure, we incorporated spatial and temporal information in the feature representation and the architecture of the neural networks. We compared Bidirectional Long Short-Term Memory Recurrent Neural Network (BLSTM-RNN)~\cite{schuster1997bidirectional}, Transformers~\cite{vaswani2017attention}, and Convolutional Neural Network (CNN)~\cite{lecun1989backpropagation}. Grid search was used for identifying hyperparameters, as well as the sliding window sizes for padding the input feature vector with preceding steps. We found that the CNN model consistently outperformed BLSTM-RNN and Transformers, which is likely due to spatial relationships of the skeleton keypoints and the relatively small size ($\approx$65k) of the handover adaptation subset of the FACT HRC dataset. Therefore, we chose the CNN model in our autonomous handover implementation. Our handover classifier includes an input layer, followed by six CNN-1D layers and one dense layer (an architecture inspired by the AlexNet~\cite{krizhevsky2017imagenet}, layer sizes [8,8,8,16,16,16,8] from bottom to top), and the output layer. The input feature vectors were padded with a window size of 5 (i.e., including 4 history time steps). The classifier was trained using the Adam optimiser~\cite{kingma2014adam} with a learning rate of 0.0001 and a batch size of 8. We adopted early stopping on validation loss with a patience of 20 epochs during training, with a maximum of 100 epochs.

Due to potential individual variances between participants and possible learning effects by the operator and the participants over time, we evaluated the model both with 5-fold CV segmented by participants (i.e., using all episodes from 20\% of participants in a testing fold) and 5-fold CV segmented by episodes (i.e., use 20\% episodes from each participant in a testing fold). Table~\ref{tab_auto_cv} reports average accuracy of the CNN classifiers on the FACT HRC dataset. 

As shown in Table~\ref{tab_auto_cv}, the timing classifiers (episode start, OTP start, OTP completion) had higher accuracy than the OTP type classifier. The difficulty in OTP type prediction is likely due to two reasons: Firstly, during teleoperation, the human operator was observed to use object layouts to identify a suitable OTP that both aligned with the participant's behaviours and reduced risks of collision with objects on the desk. As the autonomous policy only used human pose information, the reduced visual input may have limited its OTP prediction. 
Secondly, the human operator based their timing adaptation strategies mainly with short-term context (e.g., the participant had completed one assembly step and thus was ready for the next piece), while OTP location adaptation strategies involved further understanding of long-term context, such as individual preferences and task progress, which we further discuss in Section~\ref{subsec:S1vsS2_personal}. As the autonomous policy focused on key state transitions within each episode, the lack of global information may have limited its performance.


\begin{table}[tb]
\caption{Average classification accuracy on the FACT HRC dataset in 5-fold cross-validation (CV) segmented by episodes or by participants.}
\label{tab_auto_cv}
\centering
\begin{tabular}{|l|c|c|}
\hline
Classifier (\%) & Episode CV & Participant CV \\
\hline
Ep start & 98.8 & 98.2 \\
\hline
OTP start & 99.0 & 98.2 \\
\hline
OTP complete & 98.9 & 98.4 \\
\hline
OTP type & 78.6 & 74.1 \\
\hline
\end{tabular}
\end{table}

\subsection{Validating action primitives}\label{subsec:ml-primitive}
To verify our model of handover adaptation as fixed APs triggered by a set of parameterised state transitions (see Figure~\ref{episode_overview}), we investigated instances when the human operator did not follow these assumptions during teleoperated handovers in the FACT HRC dataset, i.e., when they performed timing adaptation outside of the four specified state transition points and when a non-default OTP was used. We found that the model captured 94.3\% of the operator's adaptations: in 1.2\% of episodes, the operator paused during arm reaching; in 2.3\% of episodes, the operator paused during arm tucking; in 2.8\% of episodes, the operator used non-default OTP, including stopping the end effector early. 




\FloatBarrier
\section{User Study Evaluation Results}\label{sec:s2}
We analyse the handover data, questionnaires, observation notes and interviews to evaluate the autonomous handover policy in the HRC crafting task.

\subsection{Handover episodes}\label{subsec:s2_csv}
We analysed the processed CSV data, including the observer's in-session annotation of handover type (H2R, R2H, bidirectional) and quality (good, bad, neutral), automatic log of episode count, OTP (left, middle, right of the robot), and handover timing (episode duration, pause at the work space and the storage space). Note that data from P6, P10, P14, P16 were excluded here due to incomplete annotation or episode indexing error caused by connection drops. This results in a total of 574 handover episodes from 16 participants.

\begin{table}[tb]
\caption{Handover episodes overview (mean $\pm$ std).}
\label{tab_handover_s2}
\centering
\begin{tabular}{|l|c|c|c|}
\hline
OTP & Left & Middle & Right \\
\hline
(\%) & $21.67\pm8.70$ & $78.33\pm8.70$ & $0.00$ \\
\hline
Type & R2H & H2R & Bidirectional \\
\hline
(\%) & $53.65\pm16.60$ & $31.06\pm11.58$ & $15.29\pm11.89$ \\
\hline
Quality & Good & Bad & Neutral \\
\hline
(\%) & $46.33\pm18.51$ & $37.31\pm16.00$ & $16.35\pm19.61$ \\
\hline
\hline
Timing & Ep Len & Pause (work) & Pause (store) \\
\hline
(s) & $89.42\pm21.27$ & $21.14\pm9.07$ & $25.59\pm10.22$ \\
\hline
\end{tabular}
\end{table}

As shown in Table~\ref{tab_handover_s2}, over a third of handover episodes were rated as having poor quality by the observer (i.e., bad handovers). While the predictions are binary or 3-way classifications, the diversity and subtlety in how people signalled their intent during crafting made the task challenging. We reviewed the observation notes and identified two types of errors that caused a bad episode. One type is \textit{motion primitive error} due to focusing the adaptation at key state transitions (see Figure~\ref{episode_overview}). Specifically, once object transfer completion is predicted and the robot begins to execute AP4, the autonomous model stops monitoring the participant's behaviours. Thus, when the robot is tucking its arm, it is unable to wait or react if a participant signals to the robot at this time. The other type is \textit{prediction error} due to imperfect recognition of participants' true intentions. This can be further divided into incorrect prediction in handover \textit{timing}, such as interrupting a participant's crafting activities by starting an episode early, and incorrect prediction in handover \textit{location}, such as colliding with a birdhouse kept in the middle of the work desk by performing a middle handover. 
Motion primitive error was the most common type in Stage 1 with the observer reporting 11 out of 20 participants who had at least one noticeable occurrence, while timing prediction error was the most common in Stages 2 (12 out of 20) and 3 (15 out of 20). The observation notes revealed that motion primitive error was mainly caused by participants attempting to perform bidirectional handovers instead of R2H or H2R handovers. Timing prediction error was mainly caused by the robot misrecognising a participant's assembly or painting gestures as episode or object transfer initiations, thus moving towards the participant or stretching out its arm too early and interrupting them.

\subsection{Participant behaviour and perception}\label{subsec:s2_quali}
Participants showed diverse behaviours when communicating their intention to the robot with a combination of gaze and hand gestures, similar to observations in the teleoperated sessions~\cite{tian2023crafting}. 
Recall that participants were not told if the robot was fully autonomous or being teleoperated at the start of the experiment and were asked to guess this post-study. Three participants guessed the robot was teleoperated, four were unsure, and the remaining 13 guessed correctly the robot was autonomous. In the interview, participants elaborated that the reasons for guessing the robot was teleoperated included the robot being responsive to gaze and demonstrating good collaboration that was considered to be on par with a human; The reason for being unsure was due to a mix of episodes of good collaboration and robot errors; The reasons for guessing the robot was autonomous were handover errors which they expected a human would be able to avoid. 

A one-way ANOVA test on the questionnaire responses found no significant influence of the crafting stage on participants' subjective perceptions towards the robot and the HRC.
When asked to elaborate on their impression towards the robot and the interaction in the interview, participants reported mixed responses. 
In terms of positive impressions, 11 participants reported the robot was adaptive or collaborative, with specific comments on the robot being responsive, having a good pace, being patient, or anticipating their needs by changing handover timing or location; nine participants considered the robot useful and that it performed well as a crafting assistant; seven reported the robot was natural, human-like, social, or intelligent; six reported the robot increased their satisfaction and made the crafting activity more fun or enjoyable; two commented that the robot saved physical effort for people and may help those with mobility difficulties.

In terms of negative impressions, 10 participants reported handover timing errors, including six participants reporting feeling rushed or the robot being impatient at times, and two commenting that the robot interrupted or surprised them; three reported handover location errors, including the left-handed participant; three commented the robot was inconsistent or difficult to predict; three considered having the robot as an assistant to be less efficient than crafting by themselves or the robot was being too slow; two commented that while they noticed the robot's adaptation it was not to a satisfactory performance. Compared to handover errors reported by the observer (see Section~\ref{subsec:s2_csv}), fewer participants discussed robot errors in the interview. This is consistent with~\cite{tian2023crafting}, where the participants' focus on the overall crafting experience as a user, as opposed to the operator's focus on the individual handover episodes as an observer, led to differences in their perception of handover quality.

\FloatBarrier
\section{Teleoperated vs. Auto Handover}\label{sec:S1vsS2}
\subsection{Handover episodes}\label{subsec:S1vsS2_csv}
\begin{figure}[tb]
    \centering
    \subfloat[][OTP (\%)]{\pdftooltip{\includegraphics[width=0.5\linewidth]{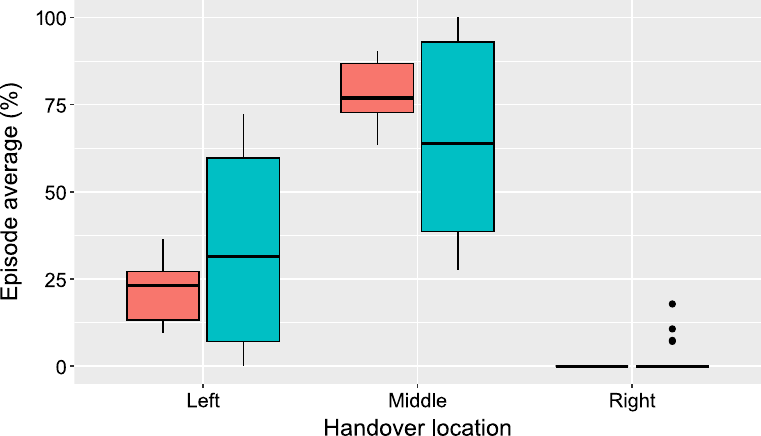}}{In this box plot, teleoperated sessions have bigger spread and higher mean in choosing left OTP, autonomous sessions have smaller spread and higher mean in choosing middle OTP, both sessions have very few right OTPs.}\label{fig:S1vS2_location}}~
    \subfloat[][Type (\%)]{\pdftooltip{\includegraphics[width=0.5\linewidth]{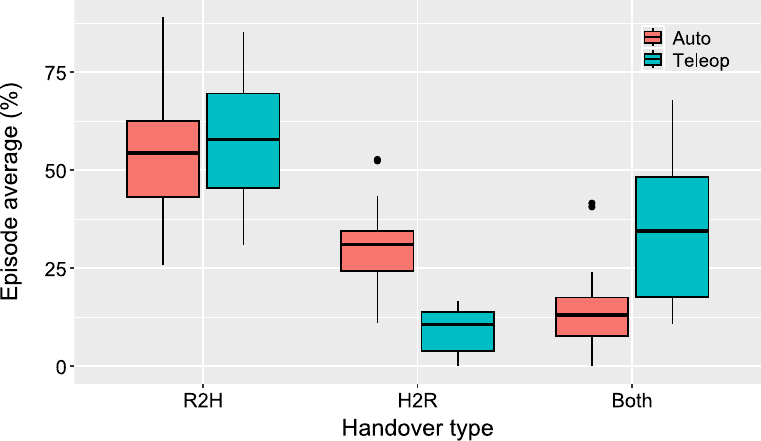}}{In this box plot, teleoperated sessions have lower mean in choosing R2H, autonomous sessions have higher mean in choosing H2R, teleoperated sessions have higher mean and wider spread in choosing bidirectional handovers.}\label{fig:S1vS2_type}}\\
    \subfloat[][Quality (\%)]{\pdftooltip{\includegraphics[width=0.5\linewidth]{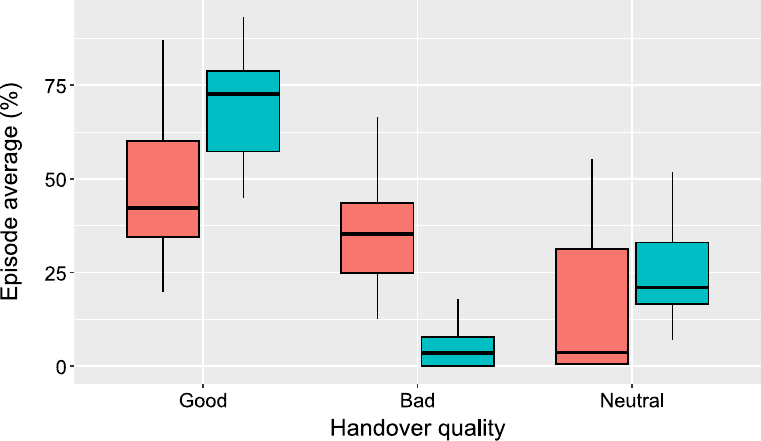}}{In this box plot, teleoperated sessions have higher mean in good handovers, autonomous sessions have higher mean in bad handovers, autonomous sessions have lower mean and wider spread in neutral handovers.}\label{fig:S1vS2_quality}}~
    \subfloat[][Timing (s)]{\pdftooltip{\includegraphics[width=0.5\linewidth]{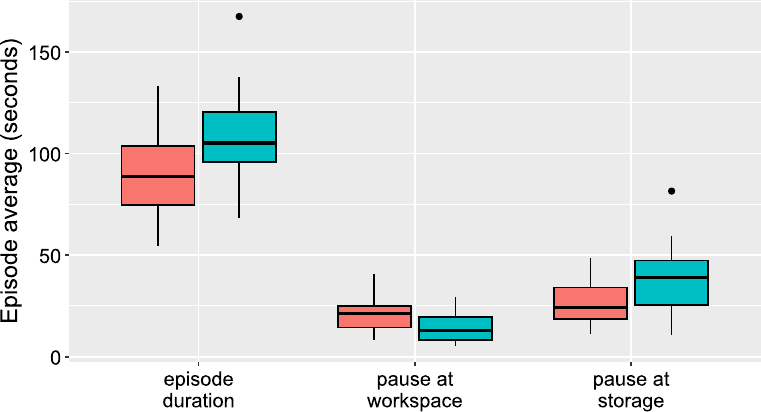}}{In this box plot, teleoperated sessions have higher mean in episode duration and pause at storage space, autonomous sessions have higher mean in pause at work space.}\label{fig:S1vS2_time}}
    \caption{Auto~vs.~teleoperated handovers in terms of OTP (a), Type (b), Quality (c) and Timing (d). More teleoperated handovers adopted left or right OTP, performed bidirectional handovers, were rated as good or neutral, with longer episodes containing longer pauses at the storage space.}
\label{fig:S1vS2_ep}
\end{figure}

We compared the handover episodes when the robot was teleoperated~\cite{tian2023crafting} and when the robot was autonomous. Note that the teleoperated CSV data included all 20 participants, while handover CSV data from P6, P10, P14, P16 in the autonomous sessions were removed (see Section~\ref{subsec:s2_csv}). We visualise the distribution of percentages of OTP location, handover type and quality, as well as handover timings when the robot was autonomous~vs.~teleoperated in Figure~\ref{fig:S1vS2_ep}.

When comparing the number of episodes, OTP location, and handover type and quality, we used episode count data and performed Bayesian AB tests with Poisson distribution to compare the two conditions. 
There are more handover episodes per participant in the autonomous sessions compared to the teleoperated sessions (probability of P(Auto$>$Tele)=99.95\%, event rate Credible Interval for interval length 0.9 is [0.10, 0.33]). Note that the teleoperated and autonomous sessions followed the same procedure with the same set of crafting objects being used. However, the human operator was more efficient with their handovers: the autonomous sessions had more single directional R2H (P(Auto$>$Tele)=98.43\%, 90\% CI [0.04, 0.33]) and H2R (P(Auto$>$Tele)=100\%, 90\% CI [1.43, 2.73]) handovers, while the teleoperated sessions had more bidirectional handovers (P(Tele$>$Audo)=100\%, 90\% CI [0.36, 0.99]). 

In terms of OTP, more autonomous episodes adopted middle OTP (P(Auto$>$Tele)=100\%, 90\% CI [0.29, 0.61]), while more teleoperated episodes adopted left (P(Tele$>$Audo)=93.37\%, 90\% CI [-0.02, 0.39]) or right (P(Tele$>$Audo)=75.54\%, 90\% CI [-0.20, 0.76]) OTP. 

Regarding handover quality, the observer rated more teleoperated handovers as good (P(Tele$>$Auto)=99.29\%, 90\% CI [0.06, 0.36]) or neutral (P(Tele$>$Auto)=85.49\%, 90\% CI [-0.07, 0.38]), while more auto handovers were rated as bad (P(Auto$>$Tele)=100\%, 90\% CI [2.64, 4.83]).

A one-way ANOVA showed that the handover model (teleoperated or autonomous) had a significant influence on average episode duration ($F=4.84$, $p=0.03$, medium effect with $\eta^{2}=0.12$). The teleoperated episodes were significantly longer on average, with shorter pauses at the workspace ($F=6.47$, $p=0.02$, large effect with $\eta^{2}=0.16$) and longer pauses at the storage space ($F=6.83$, $p=0.01$, large effect with $\eta^{2}=0.17$).

\subsection{Subjective perception}\label{subsec:S1vsS2_qualtrics}
We report participants' questionnaire responses in the teleoperated and autonomous sessions in Table~\ref{tab_qualtrics_S1vS2}. Participants reported similar impressions of the robot, except for perceived intelligence where a Wilcoxon signed rank test showed significantly higher ratings for the teleoperated session ($V = 1053$, $p=0.003$). Regarding HRC impressions, the teleoperated sessions have significantly higher ratings in fluency ($V = 1122.5$, $p=0.019$), trust ($V = 990.5$, $p=0.002$), and working alliance ($V = 1044.5$, $p=0.021$). 

\begin{table}[ht]
\begin{center}
\caption{Participants' robot and HRC impressions (mean $\pm$ std) in teleoperated vs. autonomous sessions.}
\label{tab_qualtrics_S1vS2}
\begin{tabular}{|l|c|c|}
\hline
Robot impression & Teleoperated & Autonomous \\
\hline
Anthropomorphism & $2.99\pm0.84$ & $2.95\pm0.84$ \\
\hline
Animacy & $3.14\pm0.71$ & $3.14\pm0.77$ \\
\hline
Likeability & $4.00\pm0.65$ & $3.80\pm0.61$ \\
\hline
Intelligence** & $3.70\pm0.69$ & $3.33\pm0.53$ \\
\hline
Perceived safety & $3.87\pm0.69$ & $3.89\pm0.49$ \\
\hline
\hline
HRC impression & Teleoperated & Autonomous \\
\hline
Fluency* & $3.89\pm0.60$ & $3.53\pm0.67$ \\
\hline
Trust** & $4.10\pm0.70$ & $3.54\pm0.70$ \\
\hline
Working Alliance* & $4.07\pm0.73$ & $3.76\pm0.65$ \\
\hline
Enjoyable & $4.12\pm0.58$ & $4.03\pm0.67$ \\
\hline
Satisfactory & $3.95\pm0.69$ & $3.98\pm0.62$ \\
\hline
\end{tabular}
\end{center}
\end{table}

Further, when analysing subjective ratings for each of the three crafting stages, an increasing trend from Stage 1 to Stage 3 was found when the robot employed the autonomous handover policy, albeit not significant. When the robot was teleoperated, a significant increase in subjective perception from Stage 1 to Stage 3 was found~\cite{tian2023crafting}. We hypothesise that the significant increase in subjective perception when the robot was teleoperated was caused by both increased familiarity with the robot and the crafting task, as well as the human operator's adaptation of the robot's handovers to an individual participant's working styles and preferences over time. As the autonomous robot used the same handover policy throughout the session without personalising to each individual participant, while participants had increased familiarity with the robot and the task, the lack of personalisation limited how much their perception would increase over time. We further analyse personalisation by the human operator in Section~\ref{subsec:S1vsS2_personal}.

\subsection{Personalisation in teleoperated sessions}\label{subsec:S1vsS2_personal}
\begin{figure}[tb]
    \centering
    \subfloat[][Choice of OTP (\%)]{\pdftooltip{\includegraphics[width=\linewidth]{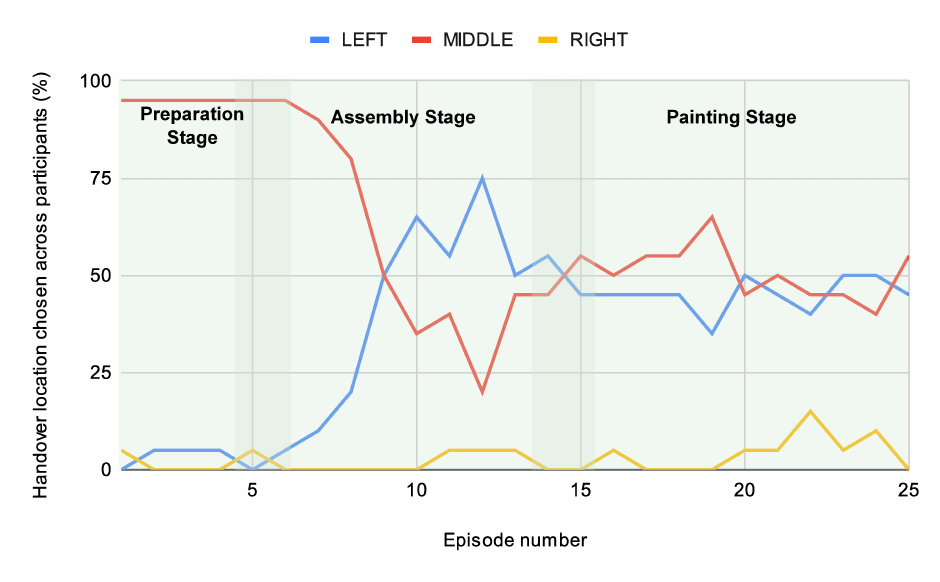}}{The x-axis is episode number, the y-axis is OTP location choice (\%). Episode 0 to 5 is approximately preparation stage, 5 to 15 is assemble stage, 15 to 25 is painting stage, with overlaps between the stages. The percentage of choice of middle handover is near 100\% at the start, but drop to around 50\% close to episode 10. The percentage of choice of left handover raised from close to 0\% to around 50\% close to episode 10. The percentage of right handovers remain low with small increases after episode 10.}\label{fig:personal_loc}}\\
    \subfloat[][Standard deviation of pause (s)]{\pdftooltip{\includegraphics[width=\linewidth]{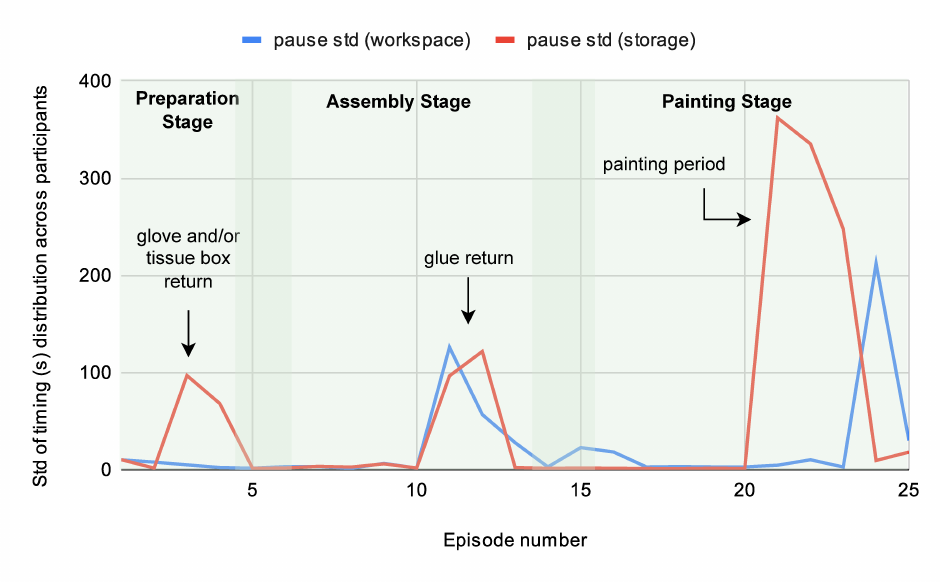}}{The x-axis is episode number, the y-axis is std of pause with a value range of 0 to 400 seconds. Episode 0 to 5 is approximately preparation stage, 5 to 15 is assemble stage, 15 to 25 is painting stage, with overlaps between the stages. The std of pause at work space showed a peak at around episode 13 with value slightly above 100, and then at around episode 24 with value slightly above 200. The std of pause at storage space showed a peak at around episode 3 and then at around episode 13 with a value slightly above 100, with a tall peak at around episode 22 with value slightly above 350.}\label{fig:personal_time}}
\caption{Handover location (a) and timing (b) choices by the operator across participants in the teleoperated HRC crafting sessions. The operator showed personalisation in their handover strategy as mixed OTPs and more diverse pauses were used in later episodes across participants.}
\label{fig:personal}
\end{figure}

We analysed the timing and location of teleoperated handovers across all 20 participants aligned by the episode ID. As shown in Fig~\ref{fig:personal_loc}, in the Preparation stage the operator chose middle OTP for most participants, while a mix of OTPs were used later. This suggests that the operator chose middle OTP as a general strategy at the start of a session then personalised the OTPs as the session progressed. Further, we examined the accuracy of the autonomous model for predicting OTPs on the FACT HRC dataset using CV by episodes (see Section~\ref{subsec:ml_picked}). As shown in Table~\ref{tab_otp_ep_per_fold}, the model yielded the highest accuracy when predicting the first 20\% of episodes in each participant (trained on the remaining 80\% of episodes) with a declining accuracy in predicting OTPs that occurred later. The increased prediction errors in later episodes indicates that the autonomous policy diverged from the human operator's behaviours more in later episodes. This supports our assumption that the operator began with a generic handover strategy but personalised it as the collaboration progressed.


\begin{table}[ht]
\caption{OTP prediction accuracy on the $n$-th 20\% of episodes in a participant on the FACT HRC dataset.}
\label{tab_otp_ep_per_fold}
\centering
\begin{tabular}{|l|c|c|c|c|c|c|}
\hline
Fold & \#1 & \#2 & \#3 & \#4 & \#5 & mean \\
\hline
Acc(\%) & 92.2 & 81.6 & 71.6 & 71.4 & 76.5 & 78.6\\
\hline
\end{tabular}
\end{table}

As shown in Fig~\ref{fig:personal_time}, the participant-wise standard deviations in pauses at the work space and storage space during an episode are higher in later episodes of the teleoperated sessions. The handover timing diverged most at three key events: when the participants returned objects to the robot at the end of the preparation and assembly stages, and in the later half of the painting stage when the participants began colouring the birdhouse. The divergence is larger in the painting stage when the participants were engaged in creative activities as opposed to functional activities of assembly. This indicates that the operator personalised the handover timing, especially when participants showed individual differences in object return preferences and in creative activities. 

\section{Discussion}\label{sec:diss}
The machine learning experiments in Section~\ref{sec:ml} illustrate that our autonomous handover model learned key temporal-spatial adaptation strategies adopted by a human operator to achieve collaborative human-robot handovers. However, the user study evaluation in Sections~\ref{sec:s2}~\&~\ref{sec:S1vsS2} showed significant gaps between autonomous handovers and teleoperation, rejecting our hypothesis that the autonomous policy will achieve human-level collaboration performance. 

As discussed in~\cite{ortenzi2021object}, a handover episode can be divided into the pre-handover phase involving perception, planning and communication, and the physical handover phase involving controls and error handling. 
During the pre-handover phase, from the robot's perspective, our analysis of handover errors in Sections~\ref{subsec:s2_csv}~\&~\ref{subsec:S1vsS2_csv} showed that it is important to incorporate both human behaviours (hand gesture, head and body movements, gaze directions) and task context (task progress, object types and locations). From the user's perspective, our analysis of user perceptions in Sections~\ref{subsec:s2_quali}~\&~\ref{subsec:S1vsS2_qualtrics} indicated that there may be a trade-off between reactive handovers (i.e., robot responding to explicit human cues) and proactive handovers (i.e., robot anticipating handovers based on past experiences and context understanding). Closer examination of the teleoperated handovers in Section~\ref{subsec:S1vsS2_personal} showed that the human operator personalised their handover strategies for individual users. Therefore, it may be beneficial for autonomous handovers to incorporate prior knowledge during the pre-handover phase alongside observations specific to that episode, in order to allow correct perception of human states and intentions, as well as correct mapping from the perceived human state to their desirable handover policy (i.e., personalisation).

During the physical handover phase, our analysis of autonomous handovers and participant behaviours in Sections~\ref{subsec:s2_csv}~\&~\ref{subsec:s2_quali} showed that autonomous handovers can benefit from increased transparency and explainability, allowing a user to understand the robot's observations, plans, and actions and thus providing explicit or implicit feedback to adjust or correct them. This allows a human-in-the-loop approach to error handling and personalisation from the robot's perspective, while increasing control and agency from the user's perspective. However, such close human supervision may also conflict with the desire towards proactive handovers or efficient HRC with less user intervention. Thus, it is important to understand how different user and task context may influence the locus of control (robot vs. user) when designing collaborative handovers.

While providing insights on collaborative handovers, this study has a number of limitations. The simplification of the handover episode into a set of representative motion primitives (see Section~\ref{subsec:ml-primitive}) has limited the flexibility of the autonomous handover policy, especially in its ability to recover from recognition errors. An alternative to the supervised  learning based approach for developing the autonomous handover policy may be behaviour cloning~\cite{wang2024genh2r}, which have shown benefits in learning complex and natural robot behaviours from human demonstrations. In addition, incorporating multimodal interaction such as speech~\cite{wang2024mosaic} will increase the human-robot communication during handovers. Further, to generalise learnings from this study set in the crafting scenario to other HRC tasks, additional context should be considered, such as the human's goals or potential risk of the activities. We recommend researchers to develop handover timing and location adaptation policies specific to their human, robot, and task context, as well as to incorporate suitable personalisation and locus of control to achieve collaborative handovers.

\section{Conclusions}\label{sec:con}
We developed an autonomous handover model using the FACT HRC dataset of teleoperated handovers~\cite{tian2023crafting}, which predicts when and where a mobile manipulator robot performs key handover actions while engaging in task-oriented handovers. By conducting user study evaluation and comparing with teleoperated handovers, we identified open challenges in developing autonomous handovers that can achieve human-level HRC performance.

\section*{Acknowledgments}
This study was funded by the ARC Future Fellowship FT200100761.


\bibliographystyle{IEEEtran}
\bibliography{HRI2025_S2}
\balance

\end{document}